\documentclass{article}

\usepackage[final,nonatbib]{nips_2016}

\usepackage[utf8]{inputenc} 
\usepackage[T1]{fontenc}    
\usepackage{url}            
\usepackage{booktabs}       
\usepackage{amsfonts}       
\usepackage{nicefrac}       
\usepackage{microtype}      

\usepackage[draft]{fixme}
\usepackage[colorlinks=true, linkcolor=black, citecolor=black,
  filecolor=black, urlcolor=black]{hyperref}
\usepackage{graphicx}
\usepackage{tikz}
\usepackage{amsmath}

\title{Can Active Memory Replace Attention?}

\author{{\L}ukasz Kaiser\\ Google Brain\\ \texttt{lukaszkaiser@google.com}
\And Samy Bengio\\ Google Brain\\ \texttt{bengio@google.com}
}

\newcommand\sigmoid{\sigma}
\newcommand\argmax{\mathrm{argmax}}

\newcommand\cgru{\ensuremath{\mathrm{CGRU}}}
\newcommand\dcgru{\ensuremath{\mathrm{CGRU}^d}}
\newcommand\sfin{s_\mathrm{fin}}
\newcommand\floor[1]{\left \lfloor{#1}\right \rfloor}

\newcommand\cgruhands[1]{
\draw (0.6, 2.9) -- (1.79, 2.9);
\draw (0.6, 2.7) -- (1.79, 2.9);

\draw (0.6, 2.9) -- (1.79, 2.7);
\draw (0.6, 2.7) -- (1.79, 2.7);
\draw (0.6, 2.5) -- (1.79, 2.7);

\draw (0.6, 2.7) -- (1.79, 2.5);
\draw (0.6, 2.5) -- (1.79, 2.5);
\draw (0.6, 2.3) -- (1.79, 2.5);

\node (cg) at (1.2, 1.5) {{#1}};

\draw (0.65, 0.3) -- (1.79, 0.1);
\draw (0.65, 0.1) -- (1.79, 0.1);

\draw (0.65, 0.3) -- (1.79, 0.5);
\draw (0.65, 0.5) -- (1.79, 0.5);
\draw (0.65, 0.7) -- (1.79, 0.5);

\draw (0.65, 0.1) -- (1.79, 0.3);
\draw (0.65, 0.3) -- (1.79, 0.3);
\draw (0.65, 0.5) -- (1.79, 0.3);
}

\newcommand\cgrupic[1]{
\draw[step=0.2] (0, 0) grid (0.6, 3.0);
\cgruhands{{#1}}
}

\begin{document}

\maketitle

\begin{abstract}
Several mechanisms to focus attention of a neural network on
selected parts of its input or memory have been used successfully
in deep learning models in recent years. Attention has improved
image classification, image captioning, speech recognition,
generative models, and learning algorithmic tasks, but it
had probably the largest impact on neural machine translation.

Recently, similar improvements have been obtained using alternative
mechanisms that do not focus on a single part of a memory
but operate on all of it in parallel, in a uniform way.
Such mechanism, which we call \emph{active memory}, improved over
attention in algorithmic tasks, image processing, and in generative modelling.

So far, however, active memory has not improved over
attention for most natural language processing tasks,
in particular for machine translation. We analyze this
shortcoming in this paper and propose an extended model of
active memory that matches existing attention models on neural
machine translation and generalizes better to longer sentences.
We investigate this model and explain why previous
active memory models did not succeed. Finally, we discuss
when active memory brings most benefits
and where attention can be a better choice.
\end{abstract}

\section{Introduction}

Recent successes of deep neural networks have spanned many
domains, from computer vision \cite{img12} to
speech recognition \cite{dahl12} and many other tasks.
In particular, sequence-to-sequence recurrent neural networks (RNNs)
with long short-term memory (LSTM) cells \cite{hochreiter1997}
have proven especially successful at natural language
processing (NLP) tasks, including machine translation
\cite{sutskever14,bahdanau2014neural,cho2014learning}.

The basic sequence-to-sequence architecture for machine translation
is composed of an RNN encoder which reads the source sentence
one token at a time and transforms it into a fixed-sized state vector.
This is followed by an RNN decoder, which generates the target sentence,
one token at a time, from the state vector.
While a pure sequence-to-sequence recurrent neural network can already
obtain good translation results \cite{sutskever14,cho2014learning},
it suffers from the fact that the whole sentence to be translated
needs to be encoded into a single fixed-size vector. This clearly
manifests itself in the degradation of translation quality
on longer sentences (see Figure~\ref{fig:len}) and
hurts even more when there is less training data \cite{KVparse15}.

\begin{figure}
\begin{center}
\begin{tikzpicture}[yscale=2.0,xscale=3.5] \small

\draw[step=0.2] (0, 0.4) grid (3.0, 0.6);
\draw[step=0.2] (-0.4, 0.4) grid (-0.2, 0.6);

\node (state) at (-0.3, 0.8) {state};
\node (mem) at (1.5, 0.8) {memory};

\draw[->] (0.5, 0.61) -- (0.5, 1.09);
\draw[->] (-0.3, 0.61) -- (0.5, 1.09);

\draw[fill=lightgray] (0, 1.09) rectangle (3.0, 1.3);
\draw[fill=darkgray] (0.4, 1.09) rectangle (0.6, 1.3);
\draw[yshift=0.1cm,step=0.2] (0, 0.99) grid (3.0, 1.2);

\node (mask) at (1.5, 1.5) {mask over memory};

\draw[->] (0.5, 1.31) -- (-0.3, 1.79);

\draw[step=0.2] (0, 1.79) grid (3.0, 2.0);
\draw[step=0.2] (-0.4, 1.79) grid (-0.2, 2.0);

\node (nstate) at (-0.3, 2.2) {new state};
\node (nmem) at (1.5, 2.2) {new memory = memory};

\path[->] (-0.3, 0.61) edge [bend left=30] (-0.3, 1.79);

\end{tikzpicture}
\end{center}
\caption{Attention model. The state vector is used to compute a probability
  distribution over memory. Weighted average of memory elements, with focus
  on one of them, is used to compute the new state.}
\label{fig:attn}
\end{figure}
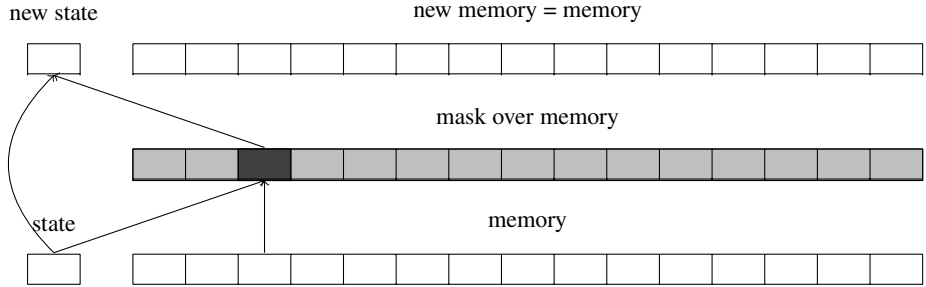

In \cite{bahdanau2014neural}, a successful mechanism to overcome
this problem was presented: a neural model of attention.
In a sequence-to-sequence model with attention, one retains
the outputs of all steps of the encoder and concatenates
them to a \emph{memory} tensor. At each step of the decoder,
a probability distribution over this memory is computed
and used to estimate a weighted average encoder representation
to be used as input to the next decoder step.
The decoder can hence focus on different parts of the encoder
representation while producing tokens. Figure~\ref{fig:attn}
illustrates a single step of this process.

The attention mechanism has proven useful well beyond the machine
translation task. Image models can benefit from attention too;
for instance, image captioning
models can focus on the relevant parts of the image when describing it~\cite{xuetal2015};
generative models for images yield especially good results with
attention, as was demonstrated by the DRAW model \cite{draw},
where the network focuses on a part of the image to produce at a given time.
Another interesting use-case for the attention mechanism is
the Neural Turing Machine \cite{ntm14}, which can learn basic
algorithms and generalize beyond the length of the training instances.

While the attention mechanism is very successful, one important
limitation is built into its definition. Since the attention mask
is computed using a Softmax, it by definition tries to focus on
a \emph{single} element of the memory it is attending to. In the
extreme case, also known as {\em hard attention}~\cite{xuetal2015},
one of the memory elements is selected and the selection is trained
using the REINFORCE algorithm  (since this is not differentiable)~\cite{reinforce}.
It is easy to demonstrate that this restriction can make some
tasks almost unlearnable for an attention model. For example,
consider the task of adding two decimal numbers, presented one
after another like this:

\begin{center}
\begin{tabular}{|c||c|c|c|c|c|c|c|c|c|}
\hline
{\bf Input}  & 1 & 2 & 5 & 0 & + & 2 & 3 & 1 & 5 \\ \hline
{\bf Output} & 3 & 5 & 6 & 5 &   &   &   &   &   \\ \hline
\end{tabular}
\end{center}

A recurrent neural network can have the carry-over in its state
and could learn to shift its attention to subsequent digits.
But that is only possible if there are \emph{two} attention heads,
attending to the first and to the second number. If only a single
attention mechanism is present, the model will have a hard time
learning this task and will not generalize properly, as was
demonstrated in~\cite{neural_gpu, stack_rnn}.

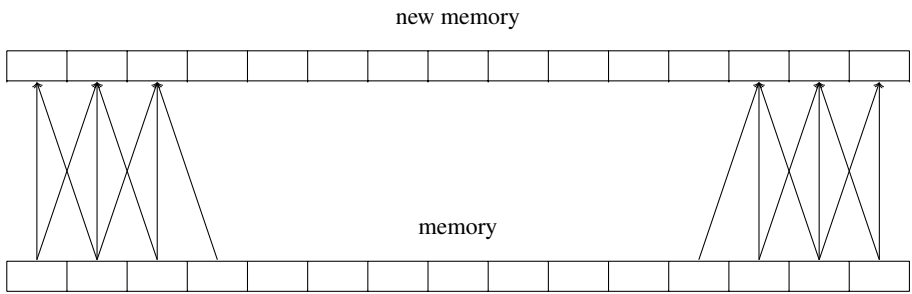
\begin{figure}[hb]
\begin{center}
\begin{tikzpicture}[yscale=2.0,xscale=4.0] \small

\draw[step=0.2] (0, 0.4) grid (3.0, 0.6);

\draw[->] (2.9, 0.61) -- (2.9, 1.79);
\draw[->] (2.7, 0.61) -- (2.9, 1.79);

\draw[->] (2.9, 0.61) -- (2.7, 1.79);
\draw[->] (2.7, 0.61) -- (2.7, 1.79);
\draw[->] (2.5, 0.61) -- (2.7, 1.79);

\draw[->] (2.7, 0.61) -- (2.5, 1.79);
\draw[->] (2.5, 0.61) -- (2.5, 1.79);
\draw[->] (2.3, 0.61) -- (2.5, 1.79);

\node (mem) at (1.5, 0.8) {memory};
\node (nmem) at (1.5, 2.2) {new memory};

\draw[->] (0.3, 0.61) -- (0.1, 1.79);
\draw[->] (0.1, 0.61) -- (0.1, 1.79);

\draw[->] (0.3, 0.61) -- (0.5, 1.79);
\draw[->] (0.5, 0.61) -- (0.5, 1.79);
\draw[->] (0.7, 0.61) -- (0.5, 1.79);

\draw[->] (0.1, 0.61) -- (0.3, 1.79);
\draw[->] (0.3, 0.61) -- (0.3, 1.79);
\draw[->] (0.5, 0.61) -- (0.3, 1.79);

\draw[step=0.2] (0, 1.79) grid (3.0, 2.0);

\end{tikzpicture}
\end{center}
\caption{Active memory model. The whole memory takes part in the computation at every
  step. Each element of memory is active and changes in a uniform way,
  e.g., using a convolution.}
\label{fig:act_mem}
\end{figure}

A solution to this problem, already proposed in the recent
literature (for instance, the Neural GPU from~\cite{neural_gpu}),
is to allow the model to access and change
all its memory at each decoding step. We will call this mechanism
an {\em active memory}. While it might seem more expensive
than attention models, it is actually not, since
the attention mechanism needs to compute an attention score for all its memory as well
in order to focus on the most appropriate part. The approximate
complexity of an attention mechanism is therefore the same as
the complexity of the active memory. In practice, we get
step-times around $1.7$ second for an active memory model,
the Extended Neural GPU introduced below, and $1.2$ second
for a comparable model with an attention mechanism.
But active memory can potentially make parallel computations
on the whole memory, as depicted in Figure~\ref{fig:act_mem}.

Active memory is a natural choice for image models as they
usually operate on a canvas. And indeed, recent works have shown
that actively updating the canvas that will be used to produce
the final results can be beneficial. Residual networks~\cite{resnet},
the currently best performing model on the ImageNet task,
falls into this category. In~\cite{poggio16} it was shown that
the weights of different layers of a residual network can be tied
(so it becomes recurrent), without degrading performance.
Other models that operate on the whole canvas at each step
were presented in~\cite{one_shot, conceptual_compression}.
Both of these models are generative and show very good performance,
yielding better results than the original DRAW model.
Thus, the active memory approach seems to be a better choice for image models.

But what about non-image models? The Neural GPUs \cite{neural_gpu} demonstrated
that active memory yields superior results on algorithmic tasks.
But can it be applied to real-world problems? In particular, the original
attention model brought a great success to natural language processing,
esp. to neural machine translation.
Can active memory be applied to this task on a large scale?

We answer this question positively, by presenting an extension of
the Neural GPU model that yields good results for neural machine translation.
This model allows us to investigate in depth a number of questions
about the relationship between attention and active memory.
We clarify why the previous active memory model did not succeed
on machine translation by showing how it is related to
the inherent dependencies in the target distributions,
and we study a few variants of the model that show
how a recurrent structure on the output side is necessary to obtain good results.

\section{Active Memory Models} \label{sec:cgrn}

In the previous section, we used the term \emph{active memory} broadly,
referring to any model where every part of the memory
undergoes active change at every step.
This is in contrast to attention models where only
a small part of the memory changes at every step,
or where the memory remains constant.

The exact implementation of an active change of the memory might
vary from model to model. In the present paper, we will focus on
the most common ways this change is implemented that all rely
on the \emph{convolution} operator.

The convolution acts on a kernel bank and a 3-dimensional tensor.
Our kernel banks are 4-dimensional tensors of shape $[k_w, k_h, m, m]$,
i.e., they contain $k_w \cdot k_h \cdot m^2$ parameters,
where $k_w$ and $k_h$ are kernel width and height. A kernel bank $U$
can be convolved with a 3-dimensional tensor $s$ of shape $[w, h, m]$ which
results in the tensor $U * s$ of the same shape as $s$ defined by:
\[ U * s[x,y,i] \ \ =\ \ \sum_{u=\floor{-k_w/2}}^{\floor{k_w/2}}\sum_{v=\floor{-k_h/2}}^{\floor{k_h/2}}\sum_{c=1}^{m}
     s[x+u,y+v,c] \cdot U[u,v,c,i]. \]

In the equation above the index $x+u$ might sometimes be negative
or larger than the size of $s$, and in such cases we assume the value
is $0$. This corresponds to the standard convolution operator used in
many deep learning toolkits, with zero padding on both sides
and stride~$1$. Using the standard operator has the advantage that
it is heavily optimized and can directly benefit from any new work
(e.g., \cite{fastconv}) on optimizing convolutions.

Given a memory tensor $s$, an active memory model will produce the next
memory $s'$ by using a number of convolutions on $s$ and combining them.
In the most basic setting, a \emph{residual} active memory model
will be defined as:
\[ s' = s + U * s, \]
i.e., it will only add to an already existing state.

While residual models have been successful in image analysis \cite{resnet} and
generation~\cite{one_shot}, they might suffer from the vanishing gradient problem
in the same way as recurrent neural networks do. Therefore, in the same spirit
as LSTM gates~\cite{hochreiter1997} and GRU gates~\cite{gru2014} improve
over pure RNNs, one can introduce convolutional LSTM and GRU operators.
Let us focus on the convolutional GRU, which we define in the same
way as in \cite{neural_gpu}, namely:
\begin{equation} \label{eq:cgru}
\begin{split}
 \cgru(s) \ &= \ u \odot s + (1 - u) \odot \tanh(U * (r \odot s) + B),
     \ \ \textrm{where} \\
 u &= \sigmoid(U' * s + B')\quad \mathrm{and}\quad r = \sigmoid(U''*s + B'').
\end{split}
\end{equation}

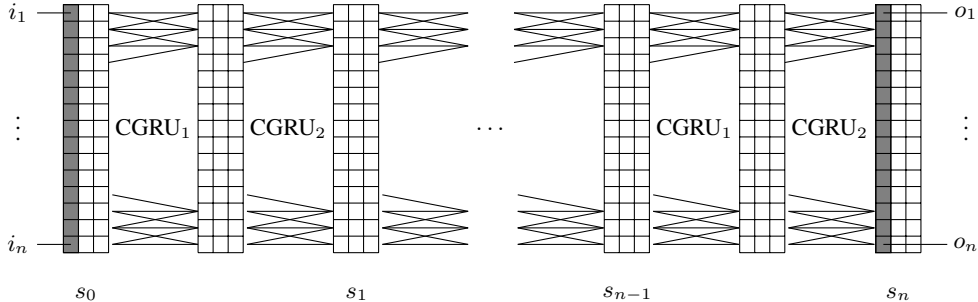
\begin{figure}
\begin{center}
\begin{tikzpicture}[yscale=1.1] \small

\draw[fill=gray] (0, 0) rectangle (0.2, 3.0);

\node(i1) at (-0.6, 2.9) {$i_1$};
\node (idots) at (-0.6, 1.6) {$\vdots$};
\foreach \y in {0.1, 2.9} {
  \draw (-0.35, \y) -- (0.1, \y);
}
\node(in) at (-0.6, 0.1) {$i_n$};

\node (s0) at (0.3, -0.5) {$s_0$};
\cgrupic{CGRU$_1$}
\begin{scope}[xshift=1.79cm]
\cgrupic{CGRU$_2$}
\end{scope}

\begin{scope}[xshift=3.59cm]
\cgrupic{}
\end{scope}
\node (s1) at (3.9, -0.5) {$s_1$};

\node (dots) at (5.7, 1.5) {$\dots$};
\begin{scope}[xshift=5.39cm]
\cgruhands{}
\end{scope}

\node (sp) at (7.5, -0.5) {$s_{n-1}$};
\begin{scope}[xshift=7.19cm]
\cgrupic{CGRU$_1$}
\begin{scope}[xshift=1.8cm]
\cgrupic{CGRU$_2$}
\end{scope}
\end{scope}

\draw[fill=gray] (10.8, 0) rectangle (11.0, 3.0);
\draw[step=0.2] (10.79, 0) grid (11.4, 3.0);
\node (sn) at (11.1, -0.5) {$s_n$};

\node(o1) at (12.0, 2.9) {$o_1$};
\node (idots) at (12.0, 1.6) {$\vdots$};
\foreach \y in {0.1, 2.9} {
  \draw (10 .9, \y) -- (11.75, \y);
}
\node(on) at (12.0, 0.1) {$o_n$};

\end{tikzpicture}
\end{center}

\caption{Neural GPU with $2$ layers and width $w=3$ unfolded in time.}
\label{fig:cgrn}
\end{figure}

As a baseline for our investigation of active memory models,
we will use the Neural GPU model from \cite{neural_gpu},
depicted in Figure~\ref{fig:cgrn}, and defined as follows.
The given sequence $i = (i_1,\ldots,i_n)$ of $n$ discrete symbols
from $\{0,\dots,I\}$ is first embedded
into the tensor $s_0$ by concatenating the vectors obtained from
an embedding lookup of the input symbols into its first column.
More precisely, we create the starting tensor $s_0$ of shape
$[w,n,m]$ by using an embedding matrix $E$ of shape $[I, m]$
and setting $s_0[0,k,:] = E[i_k]$ (in python notation)
for all $k=1 \dots n$  (here $i_1,\ldots,i_n$ is the input).
All other elements of $s_0$ are set to $0$.
Then, we apply $l$ different CGRU gates
in turn for $n$ steps to produce the final tensor $\sfin$:
\[ s_{t+1} = \cgru_l(\cgru_{l-1} \dots \cgru_1(s_t) \dots ) \quad
     \mathrm{and} \quad \sfin = s_n. \]
The result of a Neural GPU is produced by multiplying each item in
the first column of $\sfin$ by an output matrix $O$ to obtain the logits
$l_k = O \sfin[0,k,:]$ and then selecting the largest one:
$o_k = \argmax(l_k)$. During training we use the standard loss function,
i.e., we compute a Softmax over the logits $l_k$ and use
the negative log probability of the target as the loss.

\subsection{The Markovian Neural GPU}

The baseline Neural GPU model yields very poor results on neural
machine translation: its per-word perplexity on WMT\footnote{See
Section~\ref{sec:exp} for more details on the experimental setting.}
does not go below $30$ (good models on this task go below $4$),
and its BLEU scores are also very bad
(below $5$, while good models are higher than $20$).
Which part of the model is responsible for such bad results?

It turns out that the main culprit is the output generator.
As one can see in Figure~\ref{fig:cgrn} above, every output
symbol is generated independently of all other output symbols,
conditionally only on the state $\sfin$.
This is fine for learning purely deterministic functions,
like the toy tasks the Neural GPU was designed for. But it does
not work for harder real-world problems, where there could be multiple
possible outputs for each input.

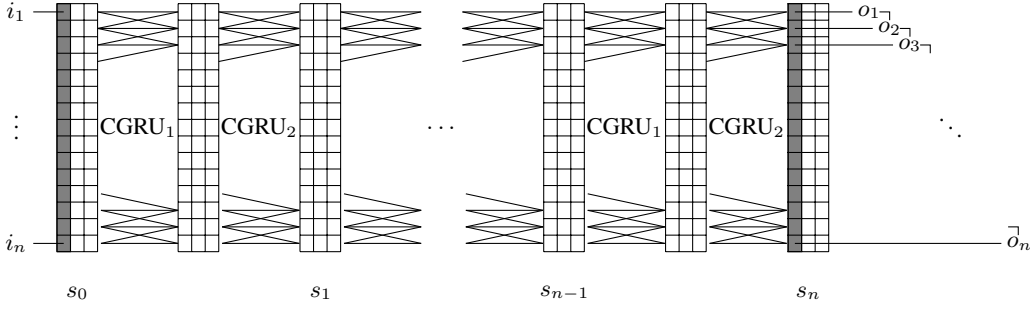
\begin{figure}
\begin{center}
\begin{tikzpicture}[yscale=1.1,xscale=0.9] \small

\draw[fill=gray] (0, 0) rectangle (0.2, 3.0);

\node(i1) at (-0.6, 2.9) {$i_1$};
\node (idots) at (-0.6, 1.6) {$\vdots$};
\foreach \y in {0.1, 2.9} {
  \draw (-0.35, \y) -- (0.1, \y);
}
\node(in) at (-0.6, 0.1) {$i_n$};

\node (s0) at (0.3, -0.5) {$s_0$};
\cgrupic{CGRU$_1$}
\begin{scope}[xshift=1.79cm]
\cgrupic{CGRU$_2$}
\end{scope}

\begin{scope}[xshift=3.59cm]
\cgrupic{}
\end{scope}
\node (s1) at (3.9, -0.5) {$s_1$};

\node (dots) at (5.7, 1.5) {$\dots$};
\begin{scope}[xshift=5.39cm]
\cgruhands{}
\end{scope}

\node (sp) at (7.5, -0.5) {$s_{n-1}$};
\begin{scope}[xshift=7.19cm]
\cgrupic{CGRU$_1$}
\begin{scope}[xshift=1.8cm]
\cgrupic{CGRU$_2$}
\end{scope}
\end{scope}

\draw[fill=gray] (10.8, 0) rectangle (11.0, 3.0);
\draw[step=0.2] (10.79, 0) grid (11.4, 3.0);
\node (sn) at (11.1, -0.5) {$s_n$};

\node(o1) at (12.0, 2.9) {$o_1$};
\node(o2) at (12.3, 2.7) {$o_2$};
\node(o3) at (12.6, 2.5) {$o_3$};
\node (idots) at (13.2, 1.6) {$\ddots$};
\node(on) at (14.22, 0.1) {$o_n$};

\draw (10.9, 2.9) -- (11.75, 2.9);
\draw (10.9, 2.7) -- (12.05, 2.7);
\draw (10.9, 2.5) -- (12.35, 2.5);
\draw (10.9, 0.1) -- (13.95, 0.1);


\foreach \i in {0, ..., 2} {
  \draw (12.15 + 0.3 * \i, 2.9 - 0.2 * \i) -- (12.3 + 0.3 * \i, 2.9 - 0.2 * \i) -- (12.3 + 0.3 * \i, 2.8 - 0.2 * \i);
}

\foreach \i in {13} {
  \draw (10.15 + 0.3 * \i, 2.9 - 0.2 * \i) -- (10.3 + 0.3 * \i, 2.9 - 0.2 * \i) -- (10.3 + 0.3 * \i, 2.8 - 0.2 * \i);
}

\end{tikzpicture}
\end{center}

\caption{Markovian Neural GPU. Each output $o_k$ is conditionally dependent
  on the final tensor $\sfin = s_n$ and the previous output symbol $o_{k-1}$.}
\label{fig:cgrnmk}
\end{figure}

The most basic way to mitigate this problem is to make every
output symbol depend on the previous output. This only changes
the output generation, not the state, so the definition of the
model is the same as above until $\sfin$. The result is then
obtained by multiplying by an output matrix $O$ each item from
the first column of $\sfin$ concatenated with the embedding of
the previous output generated by another embedding matrix $E'$:
\[ l_k = O \, \textrm{concat}(\sfin[0,k,:], E' o_{k-1}).\]
For $k=0$ we use a special symbol $o_{k-1} = \texttt{GO}$ and,
to get the output, we select $o_k = \argmax(l_k)$.
During training we use the standard loss function, i.e.,
we compute a Softmax over the logits $l_k$ and
use the negative log probability of the target as the loss.
Also, as is standard in recurrent networks~\cite{sutskever14}, we use
teacher forcing, i.e., during training we provide the true output
label as $o_{k-1}$ instead of using the previous output generated
by the model. This means that the loss incurred from generating $o_{k}$ does not directly influence the value of $o_{k-1}$.
We depict this model in Figure~\ref{fig:cgrnmk}.

\subsection{The Extended Neural GPU}

The Markovian Neural GPU yields much better results on neural
machine translation than the baseline model: its per-word
perplexity reaches about $12$ and its BLEU scores improve a bit.
But these results are still far from those achieved by models
with attention.

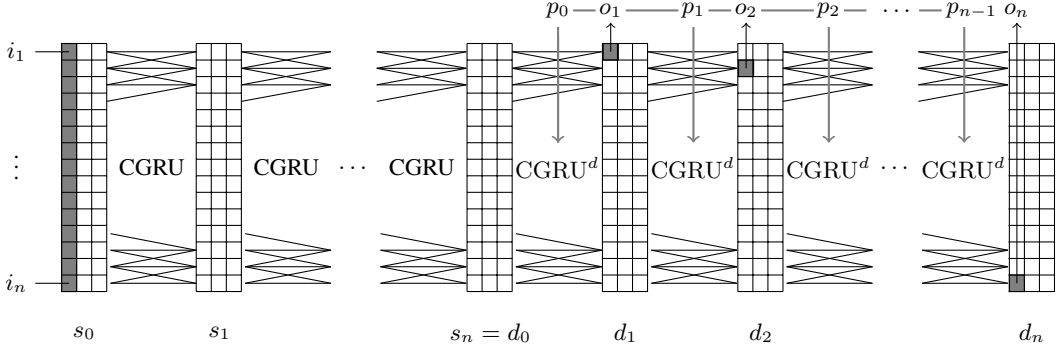
\begin{figure}
\begin{center}
\begin{tikzpicture}[yscale=1.1,xscale=1.0] \small

\draw[fill=gray] (0, 0) rectangle (0.2, 3.0);

\node(i1) at (-0.6, 2.9) {$i_1$};
\node (idots) at (-0.6, 1.6) {$\vdots$};
\foreach \y in {0.1, 2.9} {
  \draw (-0.35, \y) -- (0.1, \y);
}
\node(in) at (-0.6, 0.1) {$i_n$};

\node (s0) at (0.3, -0.5) {$s_0$};
\cgrupic{CGRU}
\begin{scope}[xshift=1.79cm]
\cgrupic{CGRU}
\end{scope}
\node (s1) at (2.1, -0.5) {$s_1$};

\begin{scope}[xshift=3.59cm]
\cgruhands{CGRU}
\end{scope}
\node (dots) at (3.9, 1.5) {$\dots$};

\begin{scope}[xshift=5.39cm]
\cgrupic{\dcgru}
\end{scope}
\node (d0) at (5.7, -0.5) {$s_n = d_0$};

\draw[fill=gray] (7.2, 3.0) rectangle (7.4, 2.8);
\begin{scope}[xshift=7.19cm]
\cgrupic{\dcgru}
\end{scope}
\node (d1) at (7.5, -0.5) {$d_1$};

\draw[thick,gray] (6.6, 3.1) -- (6.6, 1.8);

\draw[fill=gray] (9.0, 2.8) rectangle (9.2, 2.6);
\begin{scope}[xshift=8.99cm]
\cgrupic{\dcgru}
\end{scope}
\node (sp) at (9.3, -0.5) {$d_2$};

\begin{scope}[xshift=10.79cm]
\cgruhands{\dcgru}
\end{scope}
\node (dots) at (11.1, 1.5) {$\dots$};

\draw[fill=gray] (12.6, 0.2) rectangle (12.8, 0.0);
\draw[step=0.2] (12.59, 0) grid (13.2, 3.0);
\node (dn) at (12.9, -0.5) {$d_n$};

\draw[->] (7.3, 2.9) -- (7.3, 3.25);
\node(o1) at (7.3, 3.4) {$o_1$};
\draw[->] (9.1, 2.7) -- (9.1, 3.25);
\node(o2) at (9.1, 3.4) {$o_2$};
\node (dots) at (11.1, 3.4) {$\ldots$};
\draw[->] (12.7, 0.1) -- (12.7, 3.25);
\node(o2) at (12.7, 3.4) {$o_n$};

\node(p0) at (6.6, 3.4) {$p_0$};
\draw[thick,gray,->] (6.6, 3.2) -- (6.6, 1.8);
\draw[thick,gray] (6.8, 3.4) -- (7.1, 3.4);

\draw[thick,gray] (7.5, 3.4) -- (8.2, 3.4);
\node(p1) at (8.4, 3.4) {$p_1$};
\draw[thick,gray,->] (8.4, 3.2) -- (8.4, 1.8);
\draw[thick,gray] (8.6, 3.4) -- (8.9, 3.4);

\draw[thick,gray] (9.3, 3.4) -- (10.0, 3.4);
\node(p2) at (10.2, 3.4) {$p_2$};
\draw[thick,gray,->] (10.2, 3.2) -- (10.2, 1.8);
\draw[thick,gray] (10.4, 3.4) -- (10.7, 3.4);

\draw[thick,gray] (11.4, 3.4) -- (11.7, 3.4);
\node(pn) at (12.1, 3.4) {$p_{n-1}$};
\draw[thick,gray,->] (12.0, 3.2) -- (12.0, 1.8);

\end{tikzpicture}
\end{center}

\caption{Extended Neural GPU with active memory decoder. See the text
  below for definition.}
\label{fig:cgrnext}
\end{figure}

Could it be that the Markovian dependence of the outputs is too
weak for this problem, that a full recurrent dependence of the state
is needed for good performance? We test this by extending the baseline
model with an \emph{active memory decoder}, as depicted in
Figure~\ref{fig:cgrnext}.

The definition of the Extended Neural GPU follows the baseline
model until $\sfin = s_n$. We consider $s_n$ as the starting point
for the active memory decoder, i.e., we set $d_0 = s_n$. In the
active memory decoder we will also use a separate \emph{output tape tensor}
$p$ of the same shape as $d_0$, i.e., $p$ is of shape $[w,n,m]$.
We start with $p_0$ set to all $0$ and define the decoder states by
\[ d_{t+1} = \dcgru_l(\dcgru_{l-1} (\dots \dcgru_1(d_t, p_t) \dots, p_t), p_t), \]
where \dcgru is defined just like CGRU in Equation~(\ref{eq:cgru})
but with additional input as highlighted below in bold:
\begin{equation} \label{eq:dcgru}
\begin{split}
\dcgru(s, p) \ \ =\ \ u \odot s + (1 - u) \odot \tanh(U * (r \odot s) + \boldsymbol{W * p} + B),
     \ \ \textrm{where} \\
u = \sigmoid(U' * s + \boldsymbol{W' * p} + B')\quad \mathrm{and}\quad
r = \sigmoid(U''*s + \boldsymbol{W''*p} + B'').
\end{split}
\end{equation}

We generate the $k$-th output by multiplying the $k$-th vector in
the first column of $d_k$ by the output matrix $O$, i.e.,
$l_k = O \, d_k[0,k,:]$. We then select $o_k = \argmax(l_k)$.
The symbol $o_k$ is then embedded back into a dense representation
using another embedding matrix $E'$ and we put it into the $k$-th
place on the output tape $p$, i.e., we define
\[ p_{k+1} = p_k \quad \textrm{ with } \quad p_k[0,k,:] \leftarrow E' o_k. \]
In this way, we accumulate (embedded) outputs step-by-step
on the output tape $p$. Each step $p_t$ has access to all
outputs produced in all steps before $t$.

Again, it is important to note that during training we use
teacher forcing, i.e., we provide the true output labels
for $o_k$ instead of using the outputs generated by the model.

\subsection{Related Models}

A convolutional architecture has already been used to obtain good results in
word-level neural machine translation in \cite{KalchbrennerB13} and more
recently in \cite{MengLWLJL15}.
These model use a standard RNN on top of the convolution to generate the output
and avoid the output dependence problem in this way. But the state of this RNN
has a fixed size, and in the first one the sentence representation generated by
the convolutional network is also a fixed-size vector. Therefore, while superficially
similar to active memory, these models are more similar to fixed-size memory models.
The first one suffers from all the limitations of sequence-to-sequence models
without attention \cite{sutskever14,cho2014learning} that we discussed before.

Another recently introduced model, the Grid LSTM \cite{gridLSTM15},
might look less related to active memory, as it does not use convolutions
at all. But in fact it is to a large extend an active memory model -- the memory
is on the \emph{diagonal} of the grid of the running LSTM cells.
The Reencoder architecture for neural machine translation introduced in that
paper is therefore related to the Extended Neural GPU. But it differs in a number
of ways. For one, the input is provided step-wise, so the network cannot start
processing the whole input in parallel, as in our model. The diagonal memory
changes in size and the model is a 3-dimensional grid, which might not be
necessary for language processing. The Reencoder also
does not use convolutions and this is crucial for performance. The experiments
from \cite{gridLSTM15} are only performed on a very small dataset
of 44K short sentences. This is almost 1000 times smaller than the dataset
we are experimenting with and makes is unclear whether Grid LSTMs can be applied
to large-scale real-world tasks.

In image processing, in addition to the captioning \cite{xuetal2015}
and generative models \cite{one_shot, conceptual_compression} that we
mentioned before, there are several other active memory models.
They use \emph{convolutional LSTMs}, an architecture similar to CGRU,
and  have recently been used for weather prediction
\cite{convLSTMweather} and image compression \cite{convLSTMcompress},
in both cases surpassing the state-of-the-art.

\section{Experiments} \label{sec:exp}

Since all components of our models (defined above) are differentiable,
we can train them using any stochastic gradient descent optimizer.
For the results presented in this paper we used the Adam optimizer
\cite{adam} with $\varepsilon=10^{-4}$ and gradients norm clipped to $1$.
The number of layers was set to $l=2$, the width of the state tensors
was constant at $w=4$, the number of maps was $m=512$, and
the convolution kernels width and height was always $k_w=k_h=3$.%
\footnote{Our model was implemented using TensorFlow \cite{tensorflow}.
  Its code is available as open-source at
  \url{https://github.com/tensorflow/models/tree/master/neural_gpu/}.}

As our main test, we train the models discussed above and a baseline
attention model on the WMT'14 English-French translation task.
This is the same task that was used to introduce attention
\cite{bahdanau2014neural}, but -- to avoid the problem with
the \texttt{UNK} token -- we spell-out each word that is not
in the vocabulary. More precisely, we use a 32K vocabulary that
includes all characters and the most common words, and every
word that is not in the vocabulary is spelled-out letter-by-letter.
We also include a special \texttt{SPACE} symbol, which is used
to mark spaces between characters (we assume spaces between words).
We train without any data filtering on the WMT'14 corpus and
test on the WMT'14 test set (newstest'14).

As a baseline, we use a GRU model with attention that is
almost identical to the original one from \cite{bahdanau2014neural},
except that it has 2 layers of GRU cells, each with 1024 units.
Tokens from the vocabulary are embedded into vectors of size 512,
and attention is put on the top layer. This model is identical as
the one in \cite{KVparse15}, except that is uses GRU cells instead
of LSTM cells. It has about $120$M parameters, while our
Extended Neural GPU model has about $110$M parameters.
Better results have been reported on this task with attention
models with more parameters, but we aim at a baseline similar
in size to the active memory model we are using.

When decoding from the Extendend Neural GPU model, one has to provide
the expected size of the output, as it determines the size of
the memory. We test all sizes between input size and double the input
size using a greedy decoder and pick the result with smallest
log-perplexity (highest likelihood).  This is expensive, so we
only use a very basic beam-search with beam of size $2$ and no
length normalization. It is possible to reduce the cost by predicting
the output length: we tried a basic estimator based just on input
sentence length and it decreased the BLEU score by $0.3$. Better
training and decoding could remove the need to predict output length,
but we leave this for future work.

For the baseline model, we use a full beam-search
decoder with beam of size $12$, length normalization and an attention
coverage penalty in the decoder. This is a basic penalty that pushes
the decoder to attend to all words in the source sentence. We experimented
with more elaborate methods following \cite{coverage} but it did not improve
our results. The parameters for length normalization
and coverage penalty are tuned on the development set (newstest'13).
The final BLEU scores and per-word perplexities for these
different models are presented in Table~\ref{tab:res}.
Worse models have higher variance of their BLEU scores,
so we only write $< 5$ for these models.

\begin{table}
\begin{center}
\begin{tabular}{|l||c|c|}
\hline
{\bf Model}                    & {\bf Perplexity (log)} & {\bf BLEU} \\ \hline
\texttt{Neural GPU}            & 30.1 (3.5) & < 5 \\
\texttt{Markovian Neural GPU}  & 11.8 (2.5) & < 5 \\
\texttt{Extended Neural GPU}   & 3.3 (1.19) & \textbf{29.6} \\
\hline
\texttt{GRU+Attention}  & 3.4 (1.22) & 26.4 \\
\hline
\end{tabular}
\end{center}
\caption{Results on the WMT English->French translation task.
  We provide the average per-word perplexity (and its logarithm in parenthesis) and the BLEU score.
  Perplexity is computed on the test set with the ground truth provided,
  so it do not depend on the decoder.}
\label{tab:res}
\end{table}

One can see from Table~\ref{tab:res} that an active memory model
can indeed match an attention model on the machine translation task,
even with slightly fewer parameters. It is interesting to note that
the active memory model does not need the length normalization that
is necessary for the attention model (esp. when rare words are spelled).
We conjecture that active memory inherently generalizes better from shorter
examples and makes decoding easier, a welcome news, since tuning decoders
is a large problem in sequence-to-sequence models.

\begin{figure}
\begin{center}
\begin{tikzpicture}[xscale=0.17,yscale=0.5]
\draw[xstep=10cm,ystep=1cm,color=lightgray,thin] (0,15) grid (60,31);
\foreach \x in {0, 10, ..., 60} {
  \node (x) at (\x, 14.8) {\tiny \x};
}
\foreach \y in {15, 18, ..., 30} {
  \node (x) at (-2, \y) {\tiny \y};
}
\node (bottom) at (30, 14) {\small Sentence length};
\node[rotate=90] (left) at (-5, 18) {\small BLEU score};
\draw[color=blue, fill=blue] (62, 16.85) rectangle (63, 17.15);
\node[anchor=west] (l) at (63, 17)   {\tiny Extended Neural GPU};
\draw[color=red, fill=red] (62, 15.95) rectangle (63, 16.25);
\node[anchor=west] (l) at (63, 16.1) {\tiny GRU+Attention};
\draw[color=orange, fill=orange] (62, 15.05) rectangle (63, 15.35);
\node[anchor=west] (l) at (63, 15.2)   {\tiny No Attention};
\draw[thick,color=blue] plot coordinates
{ (5, 23.6)
  (15, 28.4)
  (25, 30.8)
  (35, 30.3)
  (45, 29.4)
  (55, 30.3)
};
\draw[thick,color=red,dashed] plot coordinates
{ (5, 20.6)
  (15, 24.7)
  (25, 27.7)
  (35, 27.2)
  (45, 26.4)
  (55, 23.2)
};
\draw[thick,color=orange,densely dotted] plot coordinates
{ (5, 19.7)
  (15, 22.0)
  (25, 23.0)
  (35, 21.1)
  (45, 19.2)
  (55, 14.8)
};
\end{tikzpicture}
\end{center}

\caption{BLEU score (the higher the better) vs source sentence length.}
\label{fig:len}
\end{figure}
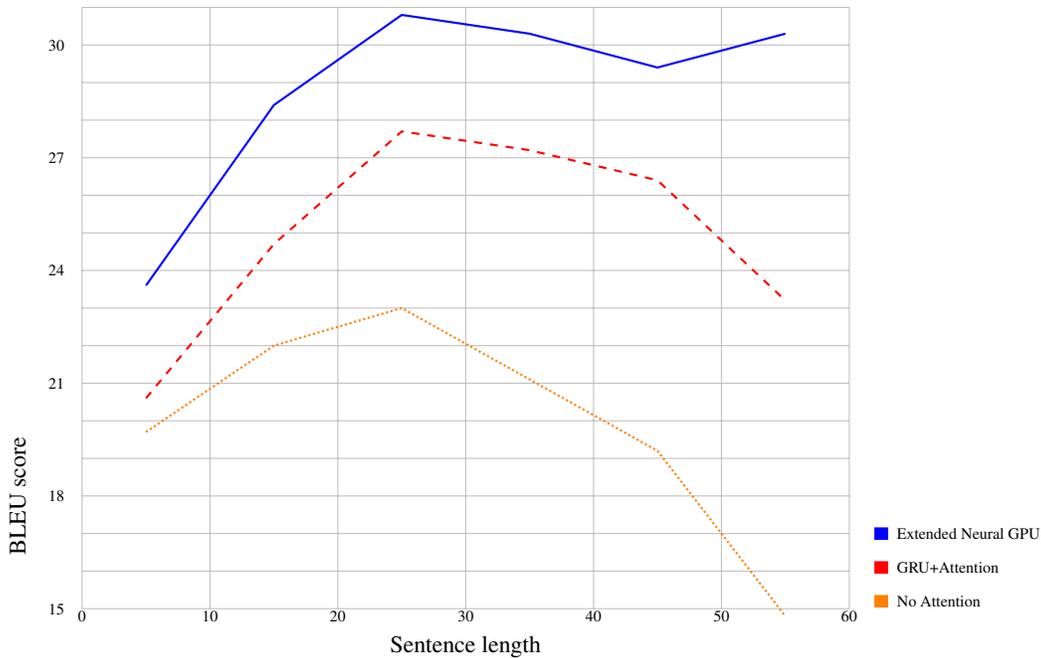

In addition to the summary results from Table~\ref{tab:res},
we analyzed the performance of the models on sentences of
different lengths. This was the key problem solved
by the attention mechanism, so it is worth asking if active memory
solves it as well.
In Figure~\ref{fig:len} we plot the BLEU scores on the test set
for sentences in each length bucket, bucketing by $10$, i.e.,
for lengths $(0, 10], (10, 20]$ and so on. We plot the curves for
the Extended Neural GPU model, the long baseline GRU model with attention,
and -- for comparison -- we add the numbers for a non-attention
model from Figure 2 of \cite{bahdanau2014neural}. (Note that
these numbers are for a model that uses different tokenization,
so they are not fully comparable, but still provide a context.)

As can be seen, our active memory model is less sensitive to sentence
length than the attention baseline. It indeed solves
the problem that the attention mechanism was  designed to solve.

\paragraph{Parsing.}
In addition to the main large-scale translation task, we tested
the Extended Neural GPU on English constituency parsing, the same
task as in \cite{KVparse15}. We only used the standard WSJ dataset
for training. It is small by neural network standards, as it contains
only 40K sentences. We trained the Extended Neural GPU with the same
settings as above, only with $m=256$ (instead of $m=512$) and
dropout of $30\%$ in each step. During decoding, we selected
well-bracketed outputs with the right number of POS-tags from
all lengths considered. Evaluated with the standard EVALB tool
on the standard WSJ 23 test set, we got $85.1$ F1 score. This is
lower than $88.3$ reported in \cite{KVparse15}, but we didn't use
any of their optimizations (no early stopping, no POS-tag substitution,
no special tuning). Since a pure sequence-to-sequence model has F1
score well below $70$, this shows that the Extended Neural GPU is
versatile and can learn and generalize well even on small data-sets.

\section{Discussion} \label{sec:discuss}

To better understand the main shortcoming of previous active memory
models, let us look at the average log-perplexities of different
attention models in Table~\ref{tab:res}. A pure Neural GPU model
yields $3.5$, a Markovian one yields $2.5$, and only a model with
full dependence, trained with teacher forcing, achieves $1.3$.
The recurrent dependence in generating the output distribution
turns out to be the key to achieving good performance.

We find it illuminating that the issue of dependencies in the output
distribution can be disentangled from the particularities of the
model or model class. In earlier works, such dependence (and training
with teacher forcing) was always used in LSTM and GRU models, but
very rarely in other kinds models. We show that it can be beneficial
to consider this issue separately from the model architecture.
It allows us to create the Extended Neural GPU and this way of
thinking might also prove fruitful for other classes of models.

When the issue of recurrent output dependencies is addressed,
as we do in the Extended Neural GPU, an active memory model can
indeed match or exceed attention models on a large-scale real-world
task. Does this mean we can always replace attention by active memory?

The answer could be \textbf{yes} for the case of soft attention.
Its cost is approximately the same as active memory, it performs
much worse on some tasks like learning algorithms, and -- with
the introduction of the Extended Neural GPU -- we do not know of a task
where it performs clearly better.

Still, an attention mask is a very natural concept, and it is probable
that some tasks can benefit from a selector that focuses on single
items by definition. This is especially obvious for hard attention:
it can be used over large memories with potentially much less computational
cost than an active memory, so it might be indispensable for
devising long-term memory mechanisms. Luckily, active memory and
attention are not exclusive, and we look forward to investigating
models that combine these mechanisms.

\small
\bibliographystyle{unsrt}
\bibliography{active_mem}

\end{document}